\documentclass[conference]{IEEEtran}
\IEEEoverridecommandlockouts
\usepackage{cite}
\usepackage{amsmath,amssymb,amsfonts}
\usepackage{algorithmic}
\usepackage{graphicx}
\usepackage{textcomp}
\usepackage{xcolor}
\usepackage{subcaption}
\usepackage{geometry}[left=54pt,right=37pt,top=105pt,bottom=54pt]
\usepackage{changepage}
\usepackage{float}

\newgeometry{left=54pt,right=37pt,top=122pt,bottom=54pt}
\def\BibTeX{{\rm B\kern-.05em{\sc i\kern-.025em b}\kern-.08em
    T\kern-.1667em\lower.7ex\hbox{E}\kern-.125emX}}

\begin{document}


\title{Towards a Unified Multidimensional Explainability Metric: Evaluating Trustworthiness in AI Models \\

}

\author
{\IEEEauthorblockN{1\textsuperscript{st} Georgios Makridis}
\IEEEauthorblockA{\textit{Department of Digital Systems} \\
\textit{University of Piraeus}\\
Piraeus, Greece \\
gmakridis@unipi.gr}
\and
\IEEEauthorblockN{2\textsuperscript{nd}Georgios Fatouros}
\IEEEauthorblockA{\textit{Department of Digital Systems} \\
\textit{University of Piraeus}\\
Piraeus, Greece \\
gfatouros@unipi.gr}
\and
\IEEEauthorblockN{3\textsuperscript{rd}Athanasios Kiourtis}
\IEEEauthorblockA{\textit{Department of Digital Systems} \\
\textit{University of Piraeus}\\
Piraeus, Greece \\
kiourtis@unipi.gr}
\and
\IEEEauthorblockN{4\textsuperscript{th} Dimitrios Kotios}
\IEEEauthorblockA{\textit{Department of Business Administration}\\
\textit{University of Piraeus}\\
Piraeus, Greece \\
dimkotios@unipi.gr}
\and
\IEEEauthorblockN{5\textsuperscript{th} Vasileios Koukos}
\IEEEauthorblockA{\textit{Department of Digital Systems} \\
\textit{University of Piraeus}\\
Piraeus, Greece \\
vkoukos@unipi.gr}
\and
\IEEEauthorblockN{6\textsuperscript{th} Dimosthenis Kyriazis}
\IEEEauthorblockA{\textit{Department of Digital Systems} \\
\textit{University of Piraeus}\\
Piraeus, Greece \\
dimos@unipi.gr}
\and
\IEEEauthorblockN{7\textsuperscript{th} Jonh Soldatos}
\IEEEauthorblockA{\textit{Innov-Acts Ltd} \\
\textit{Innov-Acts}\\
Nicosia, Cyprus\\
info@innov-acts.com}
}

\maketitle

\begin{abstract}
In this paper, we present a comprehensive framework for assessing the explainability of various XAI methods, such as LIME and SHAP, across multiple datasets and machine learning models, with the ultimate goal of creating a unified multidimensional explainability score. Our methodology focuses on three key aspects of explainability: fidelity, simplicity, and stability. We leverage benchmarking experiments to systematically evaluate these aspects and use the insights gained to construct an offline knowledge base. This knowledge base captures the explainability scores for each registered model and serves as a valuable resource for context-dependent evaluation of explainability. By analyzing the complementary characteristics and metadata of AI models, datasets, and XAI methods, the knowledge base will enable the estimation of explainability scores for previously unseen datasets and models. Properties like fidelity, simplicity, and stability may vary significantly based on the dataset, underlying model, and domain expertise of the end user. We demonstrate our framework by applying it to three open-source datasets, discussing the implications of the obtained results in relation to the characteristics of the datasets. Our work contributes to the growing field of XAI by providing a robust and versatile tool for evaluating and comparing the explainability of various XAI methods, ultimately supporting the development of more transparent and trustworthy AI systems.

\end{abstract}


\section{Introduction}
In the contemporary era, often referred to as the Digital or Information Age, intricate computational systems produce vast quantities of data on a daily basis. The digital transformation of industrial settings gives rise to the fourth industrial revolution, known as Industry 4.0 \cite{makridis2020predictive}. Artificial Intelligence (AI) serves as a critical enabler of this new era, fostering the development of cutting-edge tools and processes \cite{soldatos2021trusted}. Concurrently, there is an increasing emphasis on eXplainable Artificial Intelligence (XAI) as a means to provide intelligible explanations for the predictions and decisions generated by machine learning models.

\subsection{Motivation and Background}
XAI has emerged as a critical research area to address the growing need for interpretable and transparent AI models \cite{gilpin2018explaining} \cite{guidotti2018survey}. With the widespread adoption of AI systems across various domains, including healthcare, finance, and autonomous systems, understanding and trusting the decisions made by these systems has become paramount \cite{arrieta2020explainable}. While numerous XAI methods have been proposed to generate explanations for AI models, there still needs to be more consensus on evaluating and comparing the quality of these explanations \cite{doshi2017towards} \cite{lage2019evaluation}.

Previous research on evaluating XAI methods has focused on various properties, such as fidelity, consistency, stability, comprehensibility, and certainty \cite{ribeiro2016should} \cite{lundberg2017unified}. However, most existing evaluation frameworks are tailored to specific explanation methods or rely on subjective human judgments, which can vary based on factors like domain expertise and cognitive biases \cite{lage2019evaluation} \cite{narayanan2018humans}. Consequently, there is a pressing need for a unified multidimensional explainability metric that objectively quantifies the trustworthiness of different XAI models, facilitating their comparison and selection in real-world applications \cite{molnar1904quantifying} \cite{arrieta2020explainable}.

\subsection{Contributions and Scope}
In this paper, we propose a novel unified multidimensional explainability metric that integrates multiple aspects of XAI techniques. This metric aims to holistically capture the quality and efficiency of an AI model's trustworthiness, providing a single score to characterize the overall performance of the XAI method. 

Developing a unified multidimensional explainability metric is expected to address several challenges faced by researchers and practitioners working with AI systems. Firstly, it will help mitigate the limitations of existing evaluation frameworks, which often depend on subjective human judgments or are tailored to specific explanation methods \cite{doshi2017towards}\cite{lage2019evaluation}. By providing an objective and holistic measure of explainability, our metric will reduce the impact of domain expertise and cognitive biases on the evaluation of XAI methods \cite{narayanan2018humans}.

Secondly, our proposed metric will facilitate the comparison and selection of suitable XAI methods for different applications and domains, enabling decision-makers to make more informed choices about the trade-offs between various explanation properties \cite{arrieta2020explainable}. This is particularly important in safety-critical, such as \cite{makridis2023deep}, and regulated industries (\cite{kotios2022deep} \cite{fatouros2022deepvar}), where explainability and transparency are essential for gaining stakeholder trust and ensuring compliance with legal and ethical requirements \cite{gilpin2018explaining}.

Lastly, the proposed unified multidimensional explainability metric will contribute to the development of more interpretable and trustworthy AI models by providing researchers with a common benchmark for evaluating and improving their XAI techniques. This will encourage the development of novel explanation methods that optimize for the desired properties, ultimately leading to AI systems that are better aligned with human values and more accessible to non-expert users.

The remainder of the paper is structured as follows: Section 2 provides an overview of existing XAI methods and their evaluation frameworks, highlighting the limitations and challenges associated with current approaches to measuring explainability, while Section 3 introduces our proposed explainability framework, detailing its mathematical formulation and the rationale behind the integration of various XAI properties. Section 4 dives deeper into a preliminary experimental evaluation of our explainability metric on a range of AI models and datasets. Section 5 concludes with recommendations for future research and the potential of the current study. 


\section{Related Work}

\subsection{Explainable AI Methods}

XAI methods aim to provide human-interpretable explanations for the predictions and decisions made by machine learning models. Some of the most popular and widely-used XAI methods, classified into different categories based on their approach.

\subsubsection{Model-Agnostic Methods}

Model-agnostic methods are designed to explain any machine learning model without requiring specific knowledge about the internal workings of the model.
\begin{itemize}
    \item LIME (Local Interpretable Model-agnostic Explanations) \cite{ribeiro2016should}: LIME generates local explanations by approximating the complex model with a simpler interpretable model (e.g., linear regression) in the neighbourhood of a specific instance.
    \item SHAP (SHapley Additive exPlanations) \cite{lundberg2017unified}: SHAP values are based on game theory and provide a unified measure of feature importance for individual predictions, fairly distributing the prediction output among input features.
    \item Anchors \cite{ribeiro2018anchors}: Anchors provide rule-based explanations by identifying a minimal set of conditions (anchors) that, when satisfied, guarantee the same prediction for instances in a local region.
\end{itemize}

\subsubsection{Model-Specific Methods}

Model-specific methods provide explanations for specific types of machine learning models by exploiting their inherent structure or properties.
\begin{itemize}
    \item Decision Trees and Rule-Based Models \cite{quinlan1986induction}: These inherently interpretable models generate explanations in the form of decision rules, which can be traced along the tree structure to understand the model's predictions.
    \item Saliency Maps for Convolutional Neural Networks (CNNs) \cite{simonyan2014very}: Saliency maps highlight the most influential regions of the input image for a specific prediction, providing visual explanations for CNN predictions.
\end{itemize}

Another insightful taxonomy is the post-hoc explanation methods (this category is orthogonal to the previous 1 and 2 dimensions)
Post-hoc explanation methods involve extracting explanations from an already trained model rather than generating explanations during the model's training process.
\begin{itemize}
    \item Feature Importance \cite{breiman2001random}: Feature importance scores quantify the contribution of each input feature to the model's prediction, providing a ranking of features based on their relevance.
    \item Counterfactual Explanations \cite{arrieta2020explainable}: Counterfactual explanations provide insights into the model's decision-making process by identifying the minimal changes required in the input data to achieve a different prediction outcome.
\end{itemize}

\subsection{Related Frameworks for Explainability Evaluation}

In recent years, several off-the-self frameworks and toolkits have been proposed for applying the explainability of AI models and XAI methods (i.e., AI Explainability 360, Alibi, InterpretML). These frameworks provide guidelines and implementations for various explainability methods. However, there is still a need for an explainability metric or score that can comprehensively assess the trustworthiness and effectiveness of different XAI methods in various contexts. In our proposed methodology, we aim to address this challenge by integrating quantitative and human-grounded evaluation metrics, as well as context-dependent properties of XAI algorithms and explanation instances, to develop a robust explainability score.

In this context, several studies have focused on quantifying and evaluating explainability using different approaches. \cite{molnar1904quantifying} proposed a quantitative interpretability framework based on functional decomposition, which allowed for the evaluation and comparison of diverse models' interpretability scores, contributing to the understanding of interpretability in a more quantitative manner. In \cite{narayanan2018humans}, is presented a different approach by exploring how humans understand explanations generated by machine learning systems. Their user studies provided valuable insights into the interpretability of different explanation methods and offered a human-centered perspective on quantifying interpretability. This work emphasizes the importance of considering human cognition and perception when assessing explainability. \cite{beede2020human} focused on a specific application, evaluating the explanations provided by a deep learning system for diabetic retinopathy detection. By conducting a user study with medical professionals, they assessed the impact of explanations on user performance and trust in the system. Their findings demonstrated the effectiveness of the explanations in a real-world clinical context, underscoring the practical implications of explainability in high-stakes decision-making domains. Accordingly in \cite{pruthi2022evaluating}, is emphasized not just the generation of good explanations, but also their effective communication to users. They propose to quantify the transmitted information using methods from the domain of Information Theory, highlighting the importance of communication in the explainability process. Alongside this, there are new proposals on measures for explainable AI that go beyond traditional metrics. For instance, \cite{litman2023measures} introduces metrics that are more user-centered. These include user satisfaction, mental models, curiosity, trust, and the performance of human-AI collaborations, presenting a more comprehensive picture of explainability. These newer perspectives highlight the need for a holistic evaluation of XAI techniques, taking into account not just the technical aspects but also the communication and user-centric aspects of explainability. As part of our work, we propose a comprehensive methodology that benchmarks XAI techniques against a range of criteria, with the aim of integrating these newer perspectives in future iterations of our research.

Another important aspect of explainability evaluation is the comparison of different XAI methods. \cite{poursabzi2021manipulating} conducted a user study to compare the effects of different explanation types on users' decision-making process. They examined the influence of model-generated explanations, human-generated explanations, and contrasting cases on users' understanding, trust, and reliance on machine learning systems. Their results highlighted the nuanced relationship between explanation types and human behavior, emphasizing the need for further research in this area.

The experimental results from these studies not only contribute to the understanding of the strengths and limitations of different explanation methods but also contribute to the development of more robust and comprehensive explainability metrics for XAI models.

\subsection{Challenges in Quantifying Explainability}

Quantifying explainability is a complex task due to the multifaceted nature of explainability and the diverse range of stakeholders involved in the evaluation process. Explainability is inherently subjective, as different individuals may have different preferences and understandings of explanations. This subjectivity makes it challenging to design a universally applicable explainability metric that can accurately capture the needs and expectations of all users. A potential solution to this issue is to incorporate user feedback into the evaluation process, allowing the metric to adapt to individual preferences and requirements via active learning such as \cite{holzinger2019interactive}. As discussed earlier, various explainability metrics may be more or less relevant depending on the context and the specific goals of the XAI method. For example, fidelity and stability might vary significantly based on the dataset and the underlying model is explained, while the comprehensibility of an explanation might depend on the domain expertise of the end user. This context-dependency makes it difficult to define a unified multidimensional explainability metric that can be applied across different tasks, domains, and stakeholder groups \cite{arrieta2020explainable}. There are often trade-offs between different aspects of explainability, such as simplicity, fidelity, and coverage. For instance, a highly accurate and detailed explanation might be more complex and harder for users to understand, while a simpler explanation might sacrifice some fidelity for the sake of comprehensibility. Identifying the optimal balance between these competing metrics is a challenging task that requires careful consideration of the specific goals and constraints of the XAI method and the target user group \cite{gilpin2018explaining}.

\section{Proposed Methodology}

 \begin{figure*}[h]
\includegraphics[width=\textwidth]{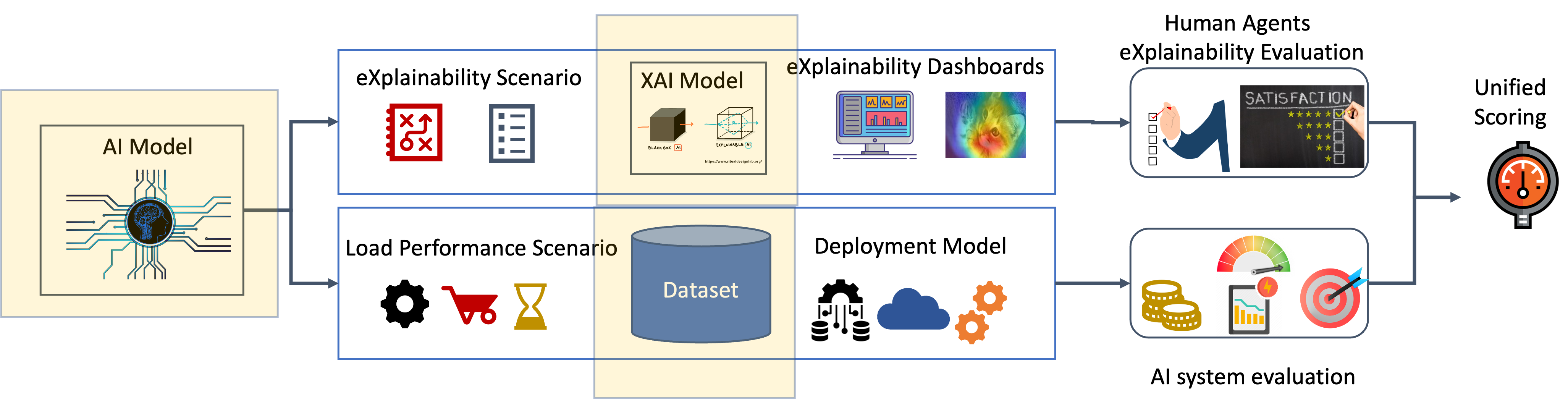}
\caption{Explainability score logical architecture with quantifying XAI and performance of the model}
\label{fig:modelxgb}
\end{figure*}

In this paper, we present our methodology for offering a framework to quantify the explainability of XAI techniques for grading the level of their trustworthiness. Our approach is based on the development of an XAI quantification scheme that factors different properties of XAI algorithms, such as fidelity, stability, simplicity, and coverage, then other properties of explanation, including user satisfaction, user trust, and user task performance are incorporated. In addition to the aforementioned requirements, we will also incorporate performance evaluation into our work, assessing the accuracy, precision, recall, and other relevant metrics of the AI models, to ensure that our explainability efforts do not compromise the effectiveness of the model's predictions as depicted in Fig.\ref{fig:modelxgb}

We first define a set of measurable properties for each component in the quantification scheme. 

\subsection{Quantitative Metrics for Explainability}

Quantitative metrics for explainability aim to provide objective assessments of the explanations generated by XAI methods. These metrics consider various properties of the explanations, such as fidelity, stability, simplicity, and coverage. In this section, we will discuss the mathematical formulations of these metrics.

\subsubsection{Fidelity}

Fidelity measures the extent to which the explanation model accurately reflects the behavior of the underlying model. A common way to compute fidelity is by using the following equation:

\begin{equation}
Fidelity = \frac{1}{N}\sum_{i=1}^N \left( \frac{\lVert f(x_i) - g(x_i) \rVert^2}{\lVert f(x_i) \rVert^2} \right),
\end{equation}

where $N$ is the number of samples, $f(x_i)$ is the prediction of the original model for sample $i$, and $g(x_i)$ is the prediction of the explanation model for the same sample.

\subsubsection{Stability}

Stability measures the consistency of explanations across different instances or minor perturbations in the input data. One way to compute stability is by using the following equation:

\begin{equation}
Stability = \frac{1}{N}\sum_{i=1}^N \frac{\lVert g(x_i) - g(x'_i) \rVert^2}{\lVert g(x_i) \rVert^2},
\end{equation}

where $x'_i$ is a perturbed version of sample $i$ and $g(x'_i)$ is the prediction of the explanation model for the perturbed sample.

\subsubsection{Average-Simplicity}

Simplicity measures the complexity of the explanations. A common way to assess simplicity is by measuring the number of features or decision rules used in the explanation. For example, for a decision tree explanation, simplicity can be computed as:

\begin{equation}
Simplicity = \frac{1}{N} \sum_{i=1}^N \frac{\text{num of nodes in the DT for sample } i}{\text{maximum num of nodes}},
\end{equation}

where $N$ is the number of samples.

\subsubsection{Coverage}

Coverage measures the proportion of instances for which the explanation method can provide an explanation to. A common way to compute coverage is by using the following equation:

\begin{equation}
Coverage = \frac{\text{in}}{N},
\end{equation}

where $N$ is the total number of instances and $in$ is the number of instances for which an explanation is provided.

User satisfaction, user trust, and user task performance are human-perceived evaluation metrics that aim to directly measure the impact of explanations on human users. These metrics can only be assessed through surveys, questionnaires, or other methods of gathering subjective opinions. 

Next, we have to assign weights to each component in the XAI quantification scheme to reflect its importance in the overall assessment. The weights can be adjusted based on the context and domain of the AI application, allowing for flexibility in prioritizing specific aspects of explainability.

The explainability score can be computed either as a weighted sum of the component values, with higher values indicating better explainability:

\begin{equation*}
  \begin{aligned}
XAI\_score = w1 * Fidelity + w2 * Stability \\ 
+ w3 * Simplicity + w4 * Coverage \\
+ w5 * User\_satisfaction + w6 * User\_trust \\
+ w7 * User\_task\_performance
\end{aligned}
\end{equation*}

where $wi$ represents the weight assigned to the ith component.

At the beginning of the benchmarking process, no specific domain is considered, so the weights will not be defined. These weights will be defined via surveys and questionnaires and will be applied in future work. 

On the other hand, for comparison of multi-dimensional objectives, aggregative weighting is not always desirable or even possible. For certain applications, it is better not to aggregate all components into single dimensions so we will offer the explainability score in a multidimensional form. 

In addition to evaluating the explainability using our proposed methodology, it may also be beneficial to contrast our explainability scores with a social performance score provided by human experts. This human-based validation methodology would involve soliciting feedback from domain experts or users who interact with the AI models and their explanations. These human evaluators would assess the quality of the explanations provided by the AI models based on their own understanding and domain knowledge.

By comparing the explainability scores generated by our methodology with the social performance scores given by humans, we can assess the extent to which our explainability metrics align with human perception and expectations. This validation approach could help us identify areas in which our quantitative explainability assessment might differ from human intuition, and refine our methodology accordingly.

 \begin{figure}[h]
\includegraphics[width=\linewidth]{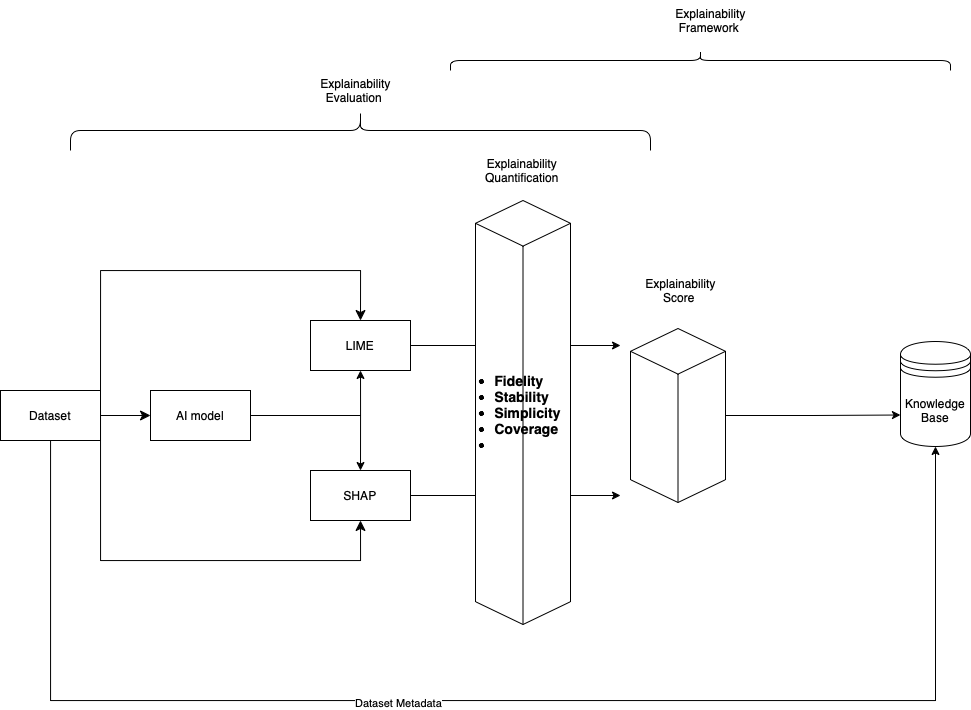}
\caption{Benchmarking experiments for Knowledge Base development}
\label{fig:banchmarking}
\end{figure}

Moreover, we envisioned that the proposed approach will be implemented as an online process, we describe our methodology for benchmarking XAI techniques and creating a knowledge base for assessing the explainability score of a variety of models. Our approach includes leveraging metadata from AI/ML models, XAI models, and datasets and evaluating various properties, such as accuracy, fidelity, consistency, stability, comprehensibility, and certainty. Furthermore, embedding of different weighing preferences, and also have in the knowledge base sensitivity of the scores to changes in these preferences. 

To validate our proposed XAI score, we applied it to a diverse set of XAI methods, such as LIME and SHAP as well as various data types, including tabular and image data

\section{Experimentation}

To begin, we performed benchmarking experiments on a diverse set of AI/ML models, XAI techniques, and datasets, encompassing various data types like tabular and image data. The purpose of these experiments is to assess the performance of different XAI methods in terms of the proposed XAI score and its components, providing a foundation for building the knowledge base.

\subsection{Datasets}

In our experiments, we employed the following datasets:

\begin{itemize}
    \item Iris dataset: This is a popular multiclass classification dataset with 150 samples, where the task is to classify iris flowers into three species (setosa, versicolor, and virginica) based on four features (sepal length, sepal width, petal length, and petal width).
    \item Wine dataset: This is another multiclass classification dataset with 178 samples, where the task is to classify wines into three classes based on 13 chemical features, such as alcohol content and color intensity.
    \item Breast Cancer Wisconsin dataset: This is a binary classification dataset with 569 samples, where the task is to classify breast cancer tumors as malignant or benign based on 30 real-valued features computed from digitized images of fine needle aspirate (FNA) of breast masses.
These datasets were chosen because they represent diverse characteristics, such as different numbers of features, sample sizes, and class distributions, which allowed us to analyze the performance of the XAI methods across various scenarios.
\end{itemize}

In each benchmarking experiment, we applied the XAI techniques to AI/ML models trained on a specific dataset and then evaluate the resulting explanations using the XAI quantification scheme. We compute the XAI score for each combination of the model, dataset, and XAI technique, considering the context-dependent nature of the properties, such as fidelity and stability, which may vary significantly depending on the dataset, and the underlying model is explained. Next, we analyze the benchmarking results to identify trends, patterns, and relationships between the properties of AI/ML models, XAI techniques, and datasets.

Based on these insights, we create a knowledge base for assessing the explainability score of various types of data assets. The knowledge base is constructed by aggregating the benchmarking results and metadata from AI/ML models, XAI models, and datasets, and then deriving rules, heuristics, or algorithms for estimating the XAI score of a given model (asset) based on its metadata.

In summary, our methodology for benchmarking XAI techniques and creating a knowledge base for assessing the explainability score of each model involves the following steps as depicted in Fig. \ref{fig:banchmarking}:

Conduct benchmarking experiments on a diverse set of AI/ML models, XAI techniques, and datasets.
Evaluate the explanations using the XAI quantification scheme and compute the XAI score for each combination of model, dataset, and XAI technique.
Analyze the benchmarking results to identify trends, patterns, and relationships between the properties of AI/ML models, XAI techniques, and datasets.
Create a knowledge base for assessing the explainability score, based on the benchmarking results and metadata from AI/ML models, XAI models, and datasets.
Periodically update the knowledge base with new benchmarking results to ensure its relevance and accuracy.

\section{Results}

We have evaluated the explainability of two popular XAI techniques, LIME and SHAP, using the benchmarking experiments and methodology. The explainability scores for LIME and SHAP were computed based on fidelity, simplicity, and stability. The results are presented in the following matrices:

\begin{table}[htbp]
\centering
\caption{LIME Explainability Scores}
\label{tab:lime_scores}
\begin{tabular}{c|c|c|c}
\hline
Dataset & Fidelity & Simplicity & Stability \\ \hline
1       & 0.4406   & 4          & 1.0000    \\
2       & 0.4397   & 13         & 0.5556    \\
3       & 0.4673   & 30         & 1.0000    \\ \hline
\end{tabular}
\end{table}

\begin{table}[htbp]
\centering
\caption{SHAP Explainability Scores}
\label{tab:shap_scores}
\begin{tabular}{c|c|c|c}
\hline
Dataset & Fidelity & Simplicity & Stability \\ \hline
1       & 7.29e-18 & 4          & 0.1385    \\
2       & 1.27e-17 & 13         & 0.0164    \\
3       & 1.23e-17 & 28.31      & 0.1521    \\ \hline
\end{tabular}
\end{table}
In order to make the result more comprehensive, in terms of evaluation we have to consider the following:
\begin{itemize}
    \item Fidelity: Range: [0, $\infty$). In our methodology, fidelity is calculated as the normalized sum of squared differences between the predictions of the original model and the explanation model across all samples. A lower fidelity score (closer to 0) indicates that the explanation model accurately represents the predictions of the original model, hence demonstrating high fidelity. Conversely, a higher fidelity score indicates that the explanation model's predictions significantly deviate from the original model's predictions, implying lower fidelity.
    \item Simplicity: Range: [1, N] where N is the total number of features in the dataset: A lower simplicity score indicates a simpler explanation, using fewer features, which is generally easier for humans to understand. A higher simplicity score, on the other hand, means that the explanation is more complex and uses more features, which may be harder for humans to comprehend
    \item Stability:Range: [-1, 1]: A high stability score (closer to 1) indicates that the XAI method generates consistent explanations for similar instances, whereas a low stability score (closer to -1) means that the explanations may vary significantly for similar instances. In general, a stable XAI method is more reliable and easier to trust, as it generates consistent explanations for similar inputs.
\end{itemize}

When evaluating the results, it is essential to consider the characteristics of the datasets and AI models used in our experiments. These characteristics include the type of data (e.g., tabular, image, text), the size of the dataset, the complexity of the underlying model, and other metadata related to the AI models, such as the training algorithms and their respective hyperparameters.

From the results, we can draw several observations, lets start with the fidelity. The fidelity scores for LIME range from 0.4397 to 0.4673, which suggests that the local models generated by LIME have a moderate fit with the global model predictions. On the other hand, the fidelity scores for SHAP are extremely close to zero (on the order of 1e-17), indicating that SHAP provides a nearly perfect fit with the global model predictions. This demonstrates that SHAP generally outperforms LIME in terms of fidelity. On the other hand, simplicity scores for both LIME and SHAP are equal, with values ranging from 4 to 30. This suggests that the complexity of the explanations generated by both methods is comparable. LIME has stability scores of either 1 or 0.5556, which indicates that its explanations are either highly stable or moderately stable across instances. In contrast, SHAP has stability scores ranging from 0.0164 to 0.1521, which are significantly lower than LIME's scores. This suggests that SHAP's explanations might be less consistent across instances compared to LIME. 

However, the above-mentioned results of the explanations may be influenced by the specific dataset and AI model, as well as the nature of the data itself. So we have to examine the results in relevance to the three datasets used. As they have different characteristics in terms of feature space, number of classes, and complexity. The differences in these characteristics can impact the explainability results obtained by LIME and SHAP.

Fidelity scores for LIME and SHAP are generally low across the three datasets, indicating that the generated explanations accurately represent the model's decision-making process. However, we can notice that the fidelity score increases as the complexity of the dataset increases. This suggests that the explanations might be less accurate for more complex datasets. Simplicity scores for both LIME and SHAP show the number of features used in the explanations. In general, as the complexity of the dataset increases, the number of features used in the explanations also increases. This indicates that the explanations generated for more complex datasets might be harder for users to understand due to the larger number of features involved. Stability scores for LIME are consistently higher than those for SHAP across the three datasets. This suggests that LIME's explanations are more consistent and less sensitive to small variations in input data. However, the stability scores for SHAP are still reasonable, indicating that SHAP's explanations are also relatively stable. It is important to note that the stability scores decrease as the complexity of the dataset increases, suggesting that the explanations might be less stable for more complex datasets.

In summary, the results reveal that SHAP outperforms LIME in terms of fidelity, while both methods have similar simplicity scores. However, LIME demonstrates higher stability compared to SHAP. These findings highlight the trade-offs between different XAI techniques, emphasizing the importance of selecting an appropriate method based on the specific requirements and context of the application, as well as the metadata of the datasets and AI models used. While in terms of the data assets used we noticed that both LIME and SHAP can provide reasonably accurate, simple, and stable explanations for the RandomForestClassifier model across different datasets. However, as the complexity of the dataset increases, the fidelity, simplicity, and stability scores might be impacted, resulting in less accurate, more complex, and less stable explanations. To better understand the explainability of LIME and SHAP, it is crucial to extend the experiments to other datasets with varying complexities, feature spaces, and class distributions.

\section{Conclusion - Future Work}

In this study, we presented a comprehensive framework for assessing and benchmarking explainability of various XAI methods, with a focus on LIME and SHAP. Our approach involved evaluating the fidelity, simplicity, and stability of these methods across multiple datasets with distinct characteristics. We demonstrated the effectiveness of our methodology by conducting benchmarking experiments and creating an offline knowledge base that leverages metadata from AI models, XAI models, and datasets to estimate explainability scores in different contexts.

Our results highlight the importance of considering the context and specific characteristics of the datasets when selecting and evaluating XAI methods. For instance, we observed that the performance of LIME and SHAP varied across datasets, indicating that their explainability scores are context-dependent. Furthermore, our analysis revealed that certain XAI techniques may be more suitable for specific types of datasets or AI models.

As part of our future work, we aim to extend our framework to include additional XAI methods and explore the incorporation of other explainability metrics. We also plan to investigate the impact of different types of AI models and their metadata on the explainability scores. 

Additionally, the scalability of our approach may be challenged by large and complex AI tasks. The computation of specific components, like fidelity and consistency, can be computationally expensive for these tasks. Future work should investigate the development of efficient approximations or sampling strategies for computing the XAI score in such cases.

\section{Acknowledgement}
The research leading to the results presented in this paper has received funding from the Europeans Union's funded Project FAME under grant agreement no 101092639.



\bibliographystyle{IEEEtran}
\bibliography{references}

\end{document}